\pdfoutput=1

\documentclass[11pt]{article}

\usepackage[preprint]{acl}

\usepackage{times}
\usepackage{latexsym}
\usepackage{algorithm, algpseudocode}
\usepackage{amsmath} 

\usepackage{booktabs, multirow, array, xcolor, colortbl, pifont}
\newcommand{\cmark}{\ding{51}}
\newcommand{\xmark}{\ding{55}}
\newcommand{\best}[1]{\textbf{#1}}
\newcommand{\uline}[1]{\underline{#1}}
\definecolor{mygreen}{rgb}{0.27,0.70,0.23}   
\definecolor{myyellow}{rgb}{0.74,0.71,0.02}  

\newcommand{\cmtg}[1]{\textcolor{mygreen}{#1}}
\newcommand{\cmty}[1]{\textcolor{myyellow}{#1}}

\definecolor{cadmiumgreen}{rgb}{0.0, 0.42, 0.24}

\definecolor{myred}{rgb}{0.7, 0.3, 0.0}
\definecolor{myblue}{rgb}{0.2, 0.3, 0.6}

\usepackage[T1]{fontenc}

\usepackage[utf8]{inputenc}

\usepackage{microtype}

\usepackage{inconsolata}

\usepackage{graphicx}

\usepackage{booktabs, multirow} 

\usepackage{tcolorbox} 

\usepackage{hyperref}

\usepackage{longtable}

\usepackage[dvipsnames]{xcolor}

\title{MAPRO: Recasting Multi-Agent Prompt Optimization as Maximum a Posteriori Inference}

\author{
    Zheyuan Zhang\textsuperscript{1}, 
    Lin Ge\textsuperscript{3},
    Hongjiang Li\textsuperscript{3},
    Weicheng Zhu\textsuperscript{3},
    Chuxu Zhang\textsuperscript{2},
    Yanfang Ye\textsuperscript{1}\textsuperscript{$\dagger$} \\
    \textsuperscript{1}University of Notre Dame, 
    \textsuperscript{2}University of Connecticut,
    \textsuperscript{3}Amazon \\
    \textsuperscript{$\dagger$}Corresponding Author \\
    \texttt{\{zzhang42, yye7\}@nd.edu},
}

\begin{document}
\maketitle
\begin{abstract}
Large language models (LLMs) have demonstrated remarkable capabilities across diverse tasks, and LLM-based agents further extend these abilities to various practical workflows. While recent progress shows that multi-agent systems (MAS) can outperform single agents by coordinating specialized roles, designing effective MAS remains difficult due to prompt sensitivity and the compounded instability MAS creates. To cope with the challenge, recent efforts in automated prompt design have reduced manual effort. However, multi-agent prompt optimization remains largely unexplored. Challenges like exponentially expanding search space and ambiguous credit assignment together make systematic design intractable without principled methods. Therefore, we introduce \textbf{M}ulti-\textbf{A}gent \textbf{PR}ompt \textbf{O}ptimization (\textbf{MAPRO}), a four-stage framework that first formulates MAS prompt optimization as a \textit{Maximum a Posteriori} (MAP) inference problem and solves it using a language-guided variant of max-product belief propagation algorithm. To address credit assignment and updates the system iteratively, MAPRO employs a topology-aware refinement mechanism that integrates execution feedback and downstream blames to selectively update agent prompts. Through this process, MAPRO progressively converges to a coordinated set of agent-specific prompt policies. Across benchmarks in various tasks, MAPRO achieves state-of-the-art performance, consistently surpassing manually engineered baselines and recent automated alternatives. Beyond performance, our MAP-based formulation also delivers general guidelines for building more reliable and principled multi-agent systems in the future \footnote{Work done during internship at Amazon.}.  
\end{abstract}

\begin{figure}[t]
	\centering
	\includegraphics[width=1\linewidth]{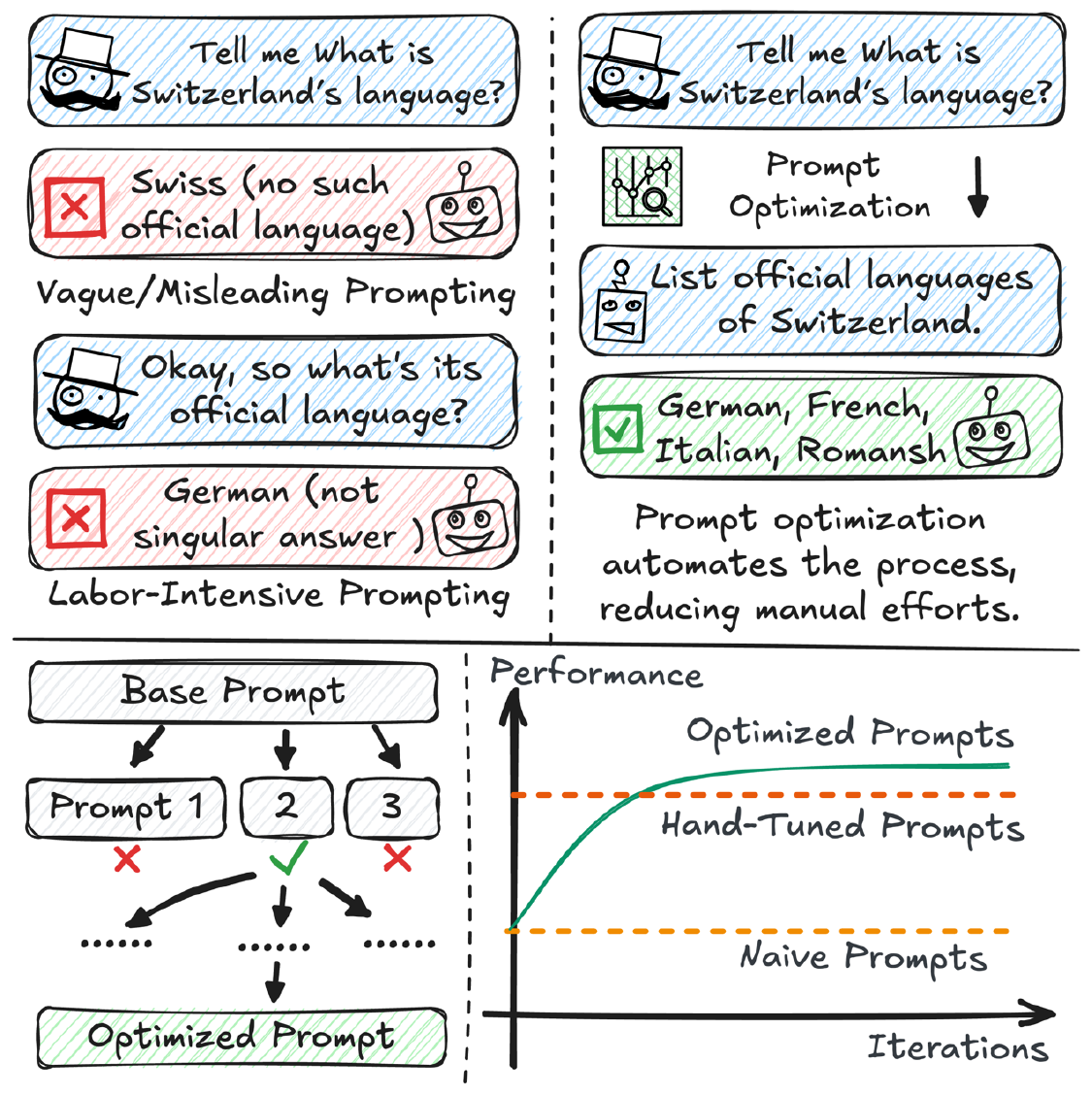}
        \vspace{-20pt}
	\caption{Prompt quality governs agent reliability: (top-left) vague, manually revised prompts are error-prone and costly; (top-right) automated prompt optimization search and produce correct answers; (bottom-left) the optimizer explores and selects among candidate rewrites; (bottom-right) performance improves over iterations, surpassing hand-tuned prompting. }
        \vspace{-25pt}
    \label{fig:motivation}
\end{figure}

\section{Introduction}
Large language models (LLMs) have emerged as powerful general-purpose learners, excelling at tasks that require reasoning, comprehension, and text generation. Their rapid progress has reshaped both research and practice across domains ranging from scientific discovery to software development \citep{kojima2022large, ouyang2022training}. Building upon this foundation, LLM-based agents have gained prominence for their ability to autonomously plan, interact, and solve complex problems with minimal human supervision. Such agents extend the reach of LLMs into practical workflows, enabling applications such as program synthesis and debugging, retrieval-augmented generation, data-centric analysis and interactive decision-making \citep{jimenez2024swe, singh2025agentic, guo2024ds, li2024embodied}. While single agents are useful, orchestrating multiple LLM agents in a coordinated system has shown even greater promise \citep{hongmetagpt, zhong2024debug, wang2024executable}. Multi-agent systems (MAS) leverage diverse perspectives and roles, such as critics, verifiers, or debaters, to collectively outperform single-agent counterparts in reasoning, exploration, and robustness \cite{shinn2023reflexion, qian2025scaling, wang2025mixture}. Yet, constructing effective MAS is far from straightforward. A recurring difficulty lies in prompt sensitivity, where small variations in instructions can drastically alter behavior and degrade performance \citep{zhoubatch}. In multi-agent settings, where outputs cascade across agents, such fragility may be compounded, amplifying instability across the system \citep{zhou2025multi}.

To mitigate these challenges, recent work has explored various forms of automated prompt design and system adaptation. Broadly, these approaches aim to reduce reliance on manual engineering by algorithmically refining prompts, adjusting agent roles, or restructuring interaction patterns \citep{khattab2024dspy, hu2025automated}. However, despite these advances, the problem of prompt optimization in multi-agent settings remains largely underexplored. This gap arises from two core challenges: (1) the search space grows combinatorially as each agent maintains its own set of prompt candidates, making it extremely difficult to navigate efficiently and leaving the system vulnerable to suboptimal local choices rather than coordinated global improvement; (2) credit assignment is highly uncertain, since it is rarely clear which agent’s prompt should be targeted for refinement, how it should be modified, or whether adjustments at the individual level will even translate into system-wide gains.

To tackle these challenges, in this paper, we propose \textbf{M}ulti-\textbf{A}gent \textbf{PR}ompt \textbf{O}ptimization (\textbf{MAPRO}), a four-stage framework that jointly explores the multi-agent prompt space, propagates feedback signals, and iteratively refines prompt policy of each agent. By grounding optimization in a principled inference process, MAPRO provides a structured approach for navigating the otherwise intractable combinatorial landscape of MAS design. Specifically, to cope with the exponential search space, we formalize multi-agent prompt optimization as a \textit{Maximum a Posteriori} (MAP) inference problem over Directed Acyclic Graphs (DAGs), and develop a language-guided variant of the max-product belief propagation (MPBP) algorithm. This design leverages agent-level and interaction-level reward models to efficiently approximate globally optimal prompt assignments in polynomial time complexity. Furthermore, to address the inherent ambiguity in credit assignment, MAPRO introduces a topology-aware refinement procedure that maintains distinct prompt policies for each agent rather than collapsing the system into a single global policy. By distributing credit by incorporating the blames from downstream agents, MAPRO progressively converging toward a set of coordinated yet agent-specific prompt policies that enhance overall system robustness and performance. Through iterative optimization, MAPRO produces multi-agent systems that achieve state-of-the-art performance, surpassing both manually engineered MAS baselines and automatically generated alternatives in single- and multi-agent settings. These improvements are consistently demonstrated across diverse tasks, including mathematical reasoning, question answering, and code generation. Our contributions can be summarized as follows: 

\begin{itemize}
    \item \textbf{MAP Inference Formulation.} To our best knowledge, we are the first to cast multi-agent prompt optimization as a \textit{Maximum a Posteriori} (MAP) inference problem. This formulation provides a principled objective for navigating the combinatorial search space, enables efficient approximation of globally optimal prompt sets, and offers general guidelines for systematic prompt optimization design.  

    \item \textbf{Topology-aware Credit Assignment.} We propose a novel refinement mechanism that integrate execution feedback and downstream blames, which alleviate the challenge of ambiguous credit assignment, enabling targeted improvements to specific agents with distinct prompt policies.  

    \item \textbf{State-of-the-Art Performance.} On diverse benchmarks—including mathematical reasoning, question answering, and code generation—MAPRO consistently surpasses manually engineered MAS baselines and recent automated alternatives, establishing new state-of-the-art results in multi-agent prompt optimization.  
\end{itemize}

\section{Preliminary}

\subsection{Multi-agent System as Directed Graph} 

We study a \emph{multi-agent system} (MAS) composed of $N$ large-language-model agents that collaborate on a shared code-generation workflow.  Let the index set of agents be $\mathcal{A}=\{1,\dots ,N\}$.  For every agent $i\in\mathcal{A}$, the agent-specific \emph{prompt candidate pool} is defined as
\begin{equation}
P_i \;=\; \bigl\{\,p_i^1,\,p_i^2,\,\dots,\,p_i^K\bigr\},
\end{equation}
where $p_i^{(k)}$ is the $k$-th candidate prompt ($k=1,\dots ,K$) and $K$ is the uniform pool size. For clarity, we denote by $p_i^*$ the optimal prompt candidate, while $\tilde{p}_i$ represents a selected prompt candidate drawn from the candidate pool. Because collaboration unfolds through directed hand-offs of textual outputs, we encode these dependencies as a directed graph $G=(\mathcal{V},\mathcal{E}), \mathcal{V}=\mathcal{A}$, in which each vertex $i\in\mathcal{V}$ corresponds to agent $i$, and a directed edge $(i,j)\in\mathcal{E}$ signifies that the output of agent $i$ is consumed as (part of) the input of agent $j$.  This graph abstraction concisely captures the information-flow topology that underpins the subsequent optimization problem.

\subsection{MAS Prompt Optimization} 

A prompt set $\tilde{P}=(\tilde{p}_1,\dots ,\tilde{p}_N)$ is considered successful if the entire workflow executed successfully and correctly.  Unlike single-agent settings, failures in MAS stem from two sources: \textbf{1) Agent Incompetence}—producing incorrect code even from well-formed input, thereby propagating errors; and \textbf{2) Defective Interaction} — an upstream agent returning semantically irrelevant text that blocks downstream progress. Both hazards need to be properly addressed to achieve good performance.

To make these notions quantitative, we define the \emph{agent score} to record the empirical quality of the $k$-th prompt of agent $i$ as $\boldsymbol{g(p_i^k)}$, and the \emph{edge score} to measure the reliability of the corresponding hand-off between the $k$-th prompt of agent $i$ and the $l$-th prompt of agent $j$ as $\boldsymbol{g(p_i^k, p_j^l)}$. Both measures lie in \([0,1]\), where the value 1 denotes flawless behavior.  To maximize the system performance and to reflect that the overall workflow is only as reliable as its weakest link, we propose the \emph{Joint Quality Score} for the multi-agent system as:
\begin{equation}
\mathcal{T}(\tilde{P})\;=\;
  \Bigl(\prod_{i=1}^{N} g(\tilde{p}_i)\Bigr)
  \Bigl(\prod_{(i,j)\in\mathcal{E}} g(\tilde{p}_i, \tilde{p}_j)\Bigr).
\end{equation}
Note that we put $\tilde{p}$ in the equation here and omit $k$ and $l$ for simplicity. Intuitively, the performance $\mathcal{T}(\tilde{P})$ of a MAS is good when every agent is competent and every hand-off is clean, because a single failure at any node or edge derails the execution. In practice, for agent $i$, there will be $K$ agent scores, and the same logic applies for the edge scores as well, so for $(i, j)$, there will be $K^2$ edge scores. Therefore, the objective of MAS prompt optimization can be defined as:
\begin{equation}
P^*\;=\;argmax_{P\in P_1\times\dots\times P_N}\mathcal{T}(P). 
\end{equation}
Equation (3) can be viewed as the \emph{posterior likelihood} that the entire system completes the evaluation batch without error, conditioned on the hidden prompt set \(P\). Indeed, if we regard each agent outcome and each edge hand-off as independent Bernoulli events given \(P\), then
under a uniform prior over prompt sets, maximizing $\boldsymbol{\mathcal{T}(P)}$ is thus equivalent to the classical \textit{maximum-a-posteriori} (MAP) problem (Proved in Appendix-\ref{appendix: proof map})

\vspace{5pt}
\noindent\textbf{Why is this problem challenging?}
\vspace{3pt}

\noindent As can be seen, the brute-force search space \(P_1\times\cdots\times P_N\) contains \(K^{N}\) discrete combinations. Moreover, the factors are highly interdependent: changing a single prompt \(p_i\) can affect many downstream agents, making greedy or local strategies prone to failure. As the objective is non-convex, discontinuous, and combinatorial, effective optimization must exploit additional structure—here, the acyclic topology of the graph \(G\) (with a prescribed iteration limit)—to prune the search space and assign credit correctly among interacting prompts. In the next section, we present an algorithm that leverages these properties to approximate \(P^*\) in polynomial time.

\begin{figure*}[htbp!]
	\centering
	\includegraphics[width=1\linewidth]{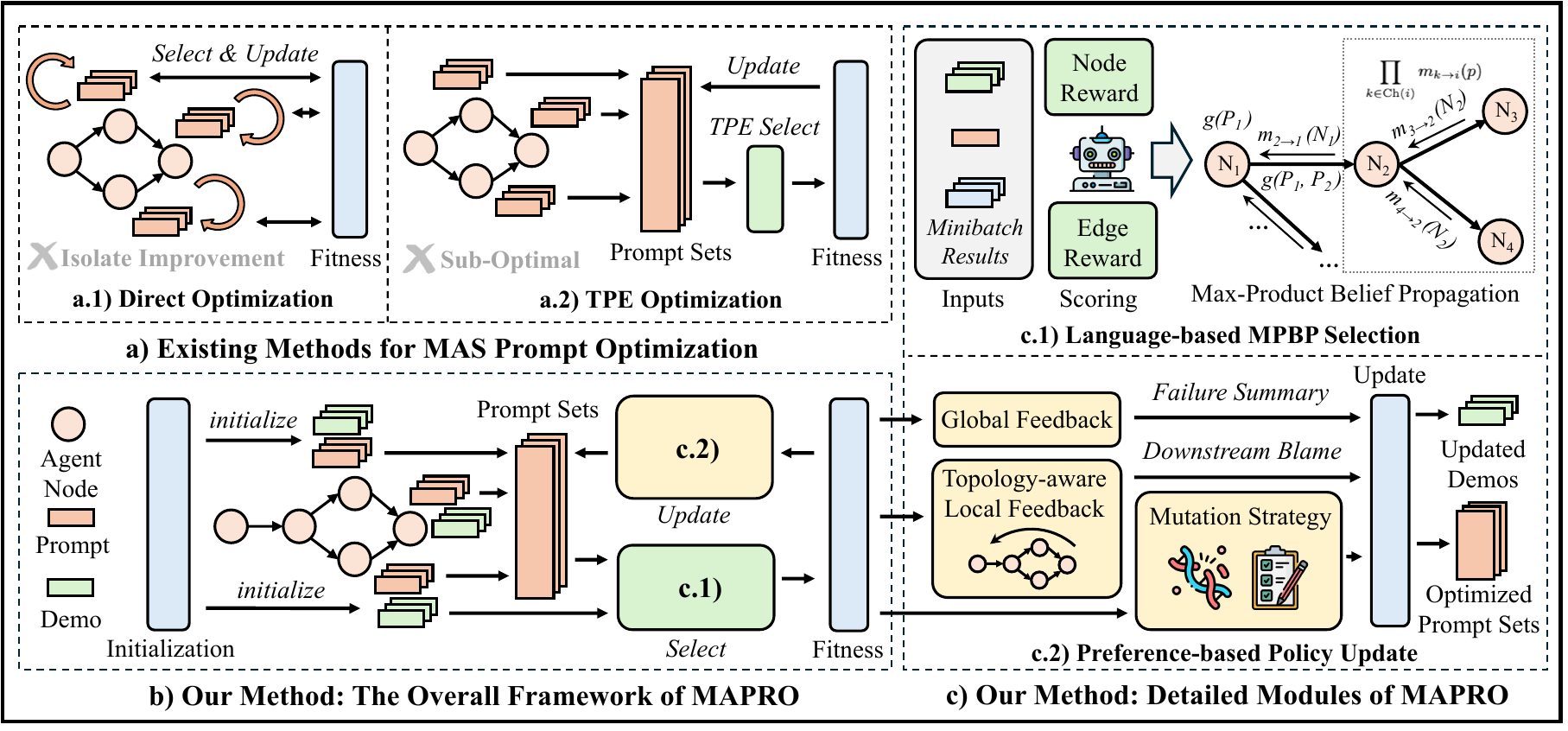}
        \vspace{-20pt}
	\caption{The Overall Framework of MAPRO. Specifically, a) shows the existing methods of prompt optimization for MAS and their drawbacks; b) shows the overall framework of MAPRO compared with existing methods; and c) demonstrate the detailed modules used in MAPRO.}
        \vspace{-20pt}
    \label{fig:framework}
\end{figure*}

\section{Methodology}

Now that we have formalized the optimization objective for multi-agent prompt optimization and highlighted the challenges, we proceed to detail our proposed framework, \textbf{M}ulti-\textbf{A}gent \textbf{PR}ompt \textbf{O}ptimization \textbf{(MAPRO)}. MAPRO addresses the Multi-agent System (MAS) prompt-optimization problem by formulating it as a discrete \emph{maximum-a-posteriori} (MAP) inference over the joint prompt space and solving it via an iterative, LLM-guided algorithm. In particular, our method comprises four stages: (1) Initialization of prompt candidates and reward models; (2) Language-based MAP selection, which employs LLM-based reward models to conduct max-product belief-propagation algorithm to efficiently find the optimal prompt combination (Figure-\ref{fig:framework} c.1); (3) Preference-based prompt policy update, which updates the prompt pools and reward models based on multi-level feedback (Figure-\ref{fig:framework} c.2); and (4) Termination, which defines stopping criteria and yields the final optimized prompt set for downstream left-out testing. We next describe each stage in detail.

\subsection{Initialization}

\paragraph{Prompt Candidate Pool Setup.}
MAPRO is designed as a plug-and-play setting atop any given MAS.  We assume an established MAS as $G=(\mathcal{V},\mathcal{E})$ (as defined in the preliminaries) and an initial set of base prompts
\(
P^0=\{p_1^0,\dots,p_N^0\}
\)
for the $N$ agents. The first step is to construct a diverse prompt candidate pool $P_i$ for each agent~$i$ by mutating its original prompt to $K$ candidates following standard practice \citep{wangpromptagent, xiang2025self}, yielding semantically similar variants
\(
P_i=\bigl\{p_i^{1},\,p_i^{2},\,\dots,\,p_i^{K}\bigr\}.
\)
 
\paragraph{Preference Demonstration Pool Setup.}
Inspired by the reward model design of TPO \citep{li2025test}, which demonstrate human preferences can be aligned during inference without retraining and achieve comparable results, we instantiate a reward model $R$ and seed the reward model with a set of \emph{accepted} (positive) and \emph{rejected} (negative) example prompt as few-shot \emph{preference demonstrations} to guide it to generate scalar scores for each agent node and edge. Intuitively, it will judge the quality of an agent's output in isolation, and the quality of a hand-off between agents. 
 
To initialize these examples, we first run the MAS on a mini-batch of training tasks $\mathcal{B}$ using a few random draws from the prompt pool $P_i$. and collect their full interaction traces if the entire task is solved correctly end-to-end on $\mathcal{B}$
Each such successful trace serves as the initial positive exemplars - the prompt $p_i$ used is considered $d^+$ and the output produced can be recorded as a \emph{desirable response} for that agent; likewise, for every edge, the message serves as an initial example of a good hand-off. To obtain complementary failure example prompts, we then generate synthetic negatives by perturbing the successful prompts as $d^-$. Thus we obtain, for each agent and edge, a pool of preference prompt responses \(\mathcal{D}=\{d^+,\, d^-\}\). The detailed input of the reward model are provided in the section below.

\subsection{Language-based MAP Selection}
Given the MAP formulation of prompt optimization from above, during the action selection phase, our goal is to efficiently find the prompt assignment $P^* = \bigl\{p_1^*,\,p_2^*,\,\dots,\,p_N^*\bigr\}$ that maximizes the joint quality score. As previously discussed, directly searching the exponentially large space $P_1 \times \cdots \times P_N$ is intractable. Therefore, we exploit the factorized structure of the multi-agent system (MAS) and introduce \textit{LLM-guided Max-Product Belief Propagation (LMPBP)}, which consists of two steps, specifically, reward model scoring and optimal prompt searching. 

\paragraph{Reward Model Scoring.} 
In the first stage, we prompt the reward model $R$ to assign numerical scores between 0 to 1. For each agent $i$, the reward model $R_i$ will rank each prompt $p_i^{k} \in P_i$ and evaluates how well the prompt would enable agent $i$ to fulfill its role. This evaluation is conditioned on the preference demonstrations $\mathcal{D}_{i}$ and the corresponding input $x_i$ and desirable response $y_i$: 
\begin{equation}
g(p_i^k) = R_i(x_i; y_i; \mathcal{D}_{i}; P_i),    
\end{equation} 
Similarly, for each directed edge $(i \to j)$, the reward model $R_{ij}$ produces a score $g(p_i^k, p_j^l)$ reflecting how well agent $i$’s output under prompt $p_i$ would set up agent $j$ for success. Concretely, we have: 
\begin{equation}
g(p_i^k, p_j^l) = R_{ij}(y_i, \mathcal{D}_{j}; P_j),    
\end{equation}
This way, we have obtained the reward scores for factors required in the searching step.  

\paragraph{Optimal Prompt Searching}

After the reward scores are secured, the second stage applies LMPBP to find the global optimum $\mathcal{T}(P)$ exactly in the DAG by passing local messages that aggregate optimal sub-solutions. For MAS with multiple parent dependencies, we convert the structure to equivalent tree-structured via a junction-tree transformation (Implementation details and equivalence proof in Appendix-\ref{appendix: proof junction tree}). The message-passing process works as follows: First, it goes through a leaf-to-root pass. (Note here the notations have different meanings) Assume agent $i$ receives messages from all of its children (downstream agents for which $i$ is an input), and then sends an aggregated message up to its own parent $j$. Specifically, for each possible prompt choice of its parent, agent $i$ computes

\begin{equation}
\label{eq:mpbp-message}
    m_{i\to j}(p_j)=\max \Bigl[g(p_i)g(p_i, p_j)\!\!\!\!\!\!\prod_{k\in\text{\textit{Child}}(i)}\!\!\!\!\!\! m_{k\to i}(p_i)\Bigr].
\end{equation}

Here $\text{\textit{Child}}(i)$ denotes the set of agents that depend on $i$’s output. Intuitively, $m_{i\to j}(p_j)$ represents the best achievable joint score of the entire subtree rooted at $i$, given that $i$’s parent $j$ is fixed to prompt $p_j$. In other words, $i$ considers all its own prompt options and those of its descendants, and encapsulates the optimal outcome (in terms of product of local scores) in a message indexed by $p_j$.  Once the upward messages reach the designated root agent $r$ (the entry point of MAS), we calculate the root belief and that agent can evaluate the total score for each of its prompt candidates using equation-\ref{eq:mpbp-message}.

This combines $r$’s own goodness score with the messages from all its children (each of which already accounts for the best configuration of that child’s subtree). We then select the highest-belief prompt for the root:
\begin{equation}
p_r^{*}\;=\;\arg\max\beta_r(p_r).
\end{equation}

Finally, we perform a downward pass to fix the prompts of the remaining agents based on the root decision. We visit each child $i$ of the root and choose the prompt that attained the maximum in $m_{i \to r}(p_r)$: 
\begin{equation}
p_i^{*}=\arg\max\Bigl[g(p_i)\,g(p_i,p_r^{*})\!\!\!\!\!\!\!\prod_{k\in\text{\textit{Child}}(i)}\!\!\!\!\!\! m_{k\to i}(p_i)\Bigr].
\end{equation}
This gives the optimal prompt for agent $i$ assuming the root was $p_r^*$. We then recursively select their best prompt given $p_i^*$, and so on, until all agents in the graph have an assigned prompt. This backtracking procedure propagates the optimal choices down the tree, yielding the globally optimal prompt set $P^*$ (Proved in Appendix-\ref{appendix: proof global}). This selected prompt set will next be used in the refinement stage to collect feedback and further improve the prompt pools and reward models.

\begin{algorithm}[t]
\caption{MAPRO Overall Process}
\begin{algorithmic}[1]

\State \textbf{Initialization:} Set up prompt pools $P$, and demonstration preferences $\mathcal{D}$.

\While{termination condition not met, }
  \State \cmtg{\textbf{// Language-based MAP Selection}}
  \State Retrieve reward scores $g(p_i)$ and $g(p_i, p_j)$.
  \State Upward pass to retrieve localized optimal 
    \Statex \hspace{2em} score $m_{i\to j}(p_j)$ at each node.
  \State Downward pass to assign best prompt $p*$ 
    \Statex \hspace{2em} given parents' choices.
  \State Run with $P^\star$ on task $\mathcal{B}$; update score $S(t)$.
  
  \State \cmty{\textbf{// Preference-guided Policy Update}}
  \State Update $\mathcal{D} \leftarrow Critic(\mathcal{D}; P; g(P))$.
  \State Get $P \leftarrow \bigl\{Mutate(\mathcal{M}(P^*), f_g, f_l), P^*\bigr\}$
  \If{improvements $\le \varepsilon$ over $T$ steps} 
        \State \textbf{break} \EndIf
  \State $t \gets t+1$
\EndWhile
\State \textbf{Inference:} Freeze $P^\star$; test on held-out tasks.

\end{algorithmic}
\end{algorithm}

\subsection{Preference-based Policy Update}
The MAP selection phase yields the global optimal prompt set $P^*$, along with an explicit assessment of each agent prompt and hand-offs via the reward scores. In the prompt policy refinement phase, we leverage this information, together with actual execution feedback or diffs on tasks, to update and improve the prompt pools and reward models. The key idea is to incorporate feedback from multiple levels: (i) global-wise execution results, (ii) downstream agent blames, and (iii) controlled prompt mutation strategy to force small edits. By integrating the multi-level feedback, we can introduce targeted prompt variations to explore new parts of the search space. After refinement, the MAS is ready to perform another round of MAP-based selection with updated components. We detail the feedback collection and update steps below.

\paragraph{Reward and Expected Output Update.}
We evaluate the performance given $P^*$ on a set of representative tasks (e.g., the training question batch $\mathcal{B}$) and we update each agent’s outputs $y_i$ as the  new \emph{desirable response} for next cycle of updates. We then use the reward scores as standards to update the accepted and rejected prompt responses for each agent. Specifically, we use a critic LLM model to judge if for agent $i$, the prompt $p_i^k$ should be updated as $d^+$ or $d^-$. Formally,  
\begin{equation}
\mathcal{D}_i \leftarrow Critic(\mathcal{D}_i; P_i; g(P_i)).
\end{equation}
This process make sure the preference demonstrations are continually updated so that the reward scores are more closely align with actual task success.

\paragraph{Prompt Pool Refinement.}
We improve the prompt candidate pool by generating new variations using a mutate LLM model with innovative feedback design from three aspects: global feedback $f_g$ indicating the final execution feedback; local feedback $f_l$ which takes the reversed topology and ask each agent to generate blames to it upstream agents based on their generated input and $f_g$, achieving fine-grained credit assignment; and a predefined mutation strategy set with small edits $\mathcal{M}$ which mimics the idea of \emph{trust region} in MAP policy optimization, keep the prompt variations from drifting afar. Formally, we invoke an LLM-based prompt mutation function to produce a refined prompt pools $P^{new}$ that modifies $P^*$: 
\begin{equation}
P_i^{new} = \bigl\{Mutate(\mathcal{M}(p_i^*), f_g, f_l), p_i^*\bigr\}. 
\end{equation}
Through such prompt augmentation, the MAS explores new prompts that are informed by past failures yet remain close to proven good prompts, thereby continuously improving robustness. Finally, the updated prompt pools are then used in the next iteration of MAP-based selection. 

\subsection{Termination}
We iterate the \textit{select--update} loop until the improvements in the joint reward have saturated, indicating convergence to an optimal prompt policy. To formalize the stopping criterion, let $S^{(t)}$ denote the joint validation score (e.g., pass rate) obtained by the best prompt set at iteration $t$. We define $\Delta S_t = S^{(t)} - S^{(t-1)}$ as the improvement in reward compared to the previous iteration. We choose a fixed \emph{patience window} size $T$ (e.g., $T=3$) and a small tolerance $\varepsilon\!\ge\!0$. After each iteration $t\!\ge\!T$, collect the recent gains
\(
\{\Delta S_{t-T+1},\dots,\Delta S_t\}.
\)
and we terminate the optimization loop when

\begin{equation}
\max_{i=1,\dots,T}\;\Delta S_{t-i+1} \;\le\; \varepsilon ,
\label{eq:stop-criterion}
\end{equation}

which means no improvement exceeding $\varepsilon$ has been observed in the last $T$ iterations. This rule halts exactly when the system has shown no progress over the specified window, ensuring that computation stops once the prompt policy has plateaued. After termination, we obtain the final optimized prompt set $P^*$ for test-time inference on unseen tasks. By locking in $P^*$, we ensure the efficiency of the system that no additional time is required during inference. The time complexity of training phase is analyzed in Appendix-\ref{appendix: time complexity}

\begin{table*}[t]
\centering
\resizebox{\textwidth}{!}{%
\begin{tabular}{c c c  c c c  c c  c c}
\toprule
\multicolumn{3}{c}{\textbf{Backbone: Claude Haiku 3.5}} &
\multicolumn{3}{c}{\textbf{Code Generation}} &
\multicolumn{2}{c}{\textbf{Question Answering}} &
\multicolumn{2}{c}{\textbf{Math Reasoning}} \\
\cmidrule(lr){1-3}\cmidrule(lr){4-6}\cmidrule(lr){7-8}\cmidrule(lr){9-10}
\textbf{Model} & \textbf{MAS} & \textbf{Optimized} &
\textbf{HumanEval-ET} & \textbf{MBPP-Plus} & \textbf{CodeContest} &
\textbf{NewsQA} & \textbf{WebQuestion} &
\textbf{MATH} & \textbf{GSM8K} \\
\midrule
Raw            & \xmark & \xmark & 69.38 $\pm$ 3.24 & 70.93 $\pm$ 0.34 & 20.36 $\pm$ 2.29 & 49.12 $\pm$ 0.11 & 33.50 $\pm$ 0.41 & 59.54 $\pm$ 1.10 & 88.57 $\pm$ 0.53 \\
CoT            & \xmark & \xmark & 70.31 $\pm$ 1.91 & 71.98 $\pm$ 0.39 & 22.91 $\pm$ 1.46 & 54.44 $\pm$ 0.30 & 33.22 $\pm$ 0.32 & 60.25 $\pm$ 0.56 & 90.71 $\pm$ 0.34 \\
ReAct          & \xmark & \xmark & 72.19 $\pm$ 1.71 & 71.02 $\pm$ 0.31 & 21.21 $\pm$ 0.61 & 58.72 $\pm$ 0.30 & 33.60 $\pm$ 0.34 & 61.29 $\pm$ 1.03 & 91.50 $\pm$ 0.39 \\
\midrule
EvoPrompt      & \xmark & \cmark & 75.63 $\pm$ 2.10 & 73.97 $\pm$ 0.48 & 22.18 $\pm$ 1.26 & 60.44 $\pm$ 0.74 & 34.99 $\pm$ 0.39 & 60.81 $\pm$ 1.66 & 92.37 $\pm$ 0.31 \\
PromptBreeder   & \xmark & \cmark & 75.31 $\pm$ 0.70 & 74.13 $\pm$ 0.22 & 21.45 $\pm$ 1.47 & 60.76 $\pm$ 0.17 & 35.12 $\pm$ 0.51 & 60.43 $\pm$ 0.52 & 92.24 $\pm$ 0.18 \\
\midrule
Chain          & \cmark & \xmark & 71.88 $\pm$ 1.10 & 74.34 $\pm$ 1.14 & 28.85 $\pm$ 2.29 & 60.88 $\pm$ 0.72 & 34.85 $\pm$ 0.40 & 62.82 $\pm$ 0.89 & 92.06 $\pm$ 0.27 \\
w/t Direct     & \cmark & \cmark & 73.96 $\pm$ 2.30 & 74.87 $\pm$ 0.53 & 29.70 $\pm$ 0.61 & 62.20 $\pm$ 1.31 & 34.25 $\pm$ 0.21 & 63.80 $\pm$ 0.21 & \uline{92.76 $\pm$ 0.14} \\
w/t TPE        & \cmark & \cmark & 75.00 $\pm$ 1.56 & \uline{75.22 $\pm$ 0.40} & 29.90 $\pm$ 0.93 & \uline{63.80 $\pm$ 0.20} & 34.01 $\pm$ 0.13 & 62.81 $\pm$ 0.37 & 92.72 $\pm$ 0.21 \\
\rowcolor{blue!8}
w/t MAPRO      & \cmark & \cmark & \best{80.21 $\pm$ 0.90} & \best{76.54 $\pm$ 0.67} & 31.52 $\pm$ 0.61 & \best{64.00 $\pm$ 0.35} & 34.65 $\pm$ 0.30 & \best{64.30 $\pm$ 0.59} & \best{93.48 $\pm$ 0.42} \\
\midrule
DMAD           & \cmark & \xmark & 72.19 $\pm$ 1.31 & 73.02 $\pm$ 0.37 & 36.77 $\pm$ 0.93 & 60.40 $\pm$ 0.37 & 34.43 $\pm$ 0.44 & 61.08 $\pm$ 0.39 & 90.39 $\pm$ 0.46 \\
w/t Direct     & \cmark & \cmark & 73.44 $\pm$ 1.56 & 74.07 $\pm$ 0.27 & 38.79 $\pm$ 0.61 & 62.20 $\pm$ 0.20 & 34.91 $\pm$ 0.07 & 62.85 $\pm$ 0.11 & 91.19 $\pm$ 0.36 \\
w/t TPE        & \cmark & \cmark & 72.92 $\pm$ 2.39 & 73.54 $\pm$ 0.53 & 37.58 $\pm$ 0.61 & 61.80 $\pm$ 0.35 & \uline{35.22 $\pm$ 0.13} & 62.81 $\pm$ 0.37 & 91.87 $\pm$ 0.29 \\
\rowcolor{blue!8}
w/t MAPRO      & \cmark & \cmark & 77.08 $\pm$ 1.81 & 74.60 $\pm$ 0.27 & 38.99 $\pm$ 1.95 & 62.93 $\pm$ 0.12 & \best{35.50 $\pm$ 0.33} & 63.33 $\pm$ 0.46 & 91.96 $\pm$ 0.58 \\
\midrule
ChatEval (Swarm)& \cmark & \xmark & 73.44 $\pm$ 1.10 & 72.60 $\pm$ 0.34 & 38.79 $\pm$ 1.05 & 60.36 $\pm$ 0.26 & 33.33 $\pm$ 0.34 & 62.62 $\pm$ 0.97 & 91.59 $\pm$ 0.89 \\
w/t Direct     & \cmark & \cmark & 74.48 $\pm$ 2.38 & 73.19 $\pm$ 0.46 & 38.18 $\pm$ 0.61 & 61.80 $\pm$ 0.20 & 34.17 $\pm$ 0.29 & \uline{63.83 $\pm$ 0.84} & 91.42 $\pm$ 0.47 \\
w/t TPE        & \cmark & \cmark & 76.04 $\pm$ 0.90 & 73.28 $\pm$ 0.26 & \uline{40.61 $\pm$ 0.61} & 62.53 $\pm$ 0.31 & 34.35 $\pm$ 0.18 & 62.68 $\pm$ 0.58 & 91.31 $\pm$ 0.62 \\
\rowcolor{blue!8}
w/t MAPRO      & \cmark & \cmark & \uline{78.13 $\pm$ 1.56} & 73.98 $\pm$ 0.15 & \best{41.41 $\pm$ 0.93} & 62.67 $\pm$ 0.31 & 34.52 $\pm$ 0.20 & 63.13 $\pm$ 0.84 & 91.73 $\pm$ 0.41 \\
\bottomrule
\end{tabular}%
}
\vspace{-5pt}
\caption{Performance results with baseline methods on Claude Haiku 3.5. We report the mean and standard deviation for all results. The best performance is bolded and runner-ups are underlined. }
\label{tab:result_haiku}
\vspace{-15pt}
\end{table*}

\begin{table}[t]
    \centering
    \renewcommand{\arraystretch}{0.9}
    \begin{tabular}{p{0.95\linewidth}}
        \toprule
        \small\textbf{Prompt Optimization Example}\\
        \midrule
        \small\textbf{Base instruction:}\\
        \small You are a Python programmer. Write pure, runnable Python code that solves the task. \\
        \midrule
        \small \cmtg{\textbf{Adding: }}\\
        \small You are a Python programmer. Write pure, runnable Python code that solves the task. 
        \cmtg{\textbf{Ensure the solution is a single function named solution}} \cmty{\textbf{with robust input validation}}, direct implementation, and \cmtg{\textbf{no type hints}}. \cmty{\textbf{Handle edge cases}} explicitly and provide clear, executable code. \\
        \midrule
        \small\cmty{\textbf{Replacement: }} \\
        \small You are a Python programmer. Write pure, runnable Python code that solves the task. Ensure the solution is a single function named solution with 
        \cmty{\textbf{robust input validation using isinstance() checks, type conversion fallbacks}}, and comprehensive error handling. \cmty{\textbf{Use try-except blocks with specific exception types, provide default values for edge cases}}, \dots \\
        \bottomrule
    \end{tabular}
    \vspace{-5pt}
    \caption{"Adding" appends guidance to the prompt while "Replacement" rewrites previous parts with better instructions. Colors highlight the modified texts.}
    \vspace{-15pt}
    \label{tab:prompt-editing}
\end{table}

\section{Experiments}
\subsection{Experimental Setup}
\noindent \textbf{Benchmarks. }We conduct experiments on an extensive collection of
tasks: HumanEval-ET, MBPP-Plus and CodeContest for code generation task, NewsQA and WebQuesion for question answering task, and MATH and GSK8K for math reasoning task. Since we are focusing on the prompt optimization and have a training scheme, we discuss the split of sets with other details including citations of these benchmarks in Appendix-\ref{appendix:Benchmarks}.

\noindent \textbf{Baselines. }We consider the following types of baselines: 1) Single agents without prompt optimization, including the raw model, and the most classical baselines CoT and ReAct; 2) Single agents with prompt optimization, including two most recent SOTA baselines EvoPrompt, and PromptBreeder. 3) Classical Multi-agent baselines without prompt optimization. While there are many MAS, we hope to choose the ones that are designed for general tasks, recent SOTA and covering as many types of common topologies as possible, therefore we choose Chain Design, DMAD, and the "swarm" version of ChatEval, which is Simultaneous-Talk-with-Summarizer. 4) Prompt optimization for MAS baselines. Since we only consider prompt optimization\footnote{In this paper, we consider the plug-and-play settings an unique advantage and don't update the topology because this is more friendly to industry scenarios where teams already have their developed MAS implemented and in production.}, we adopt the direct optimization method from GPTSwarm and Tree-structured Parzen Estimator (TPE) methods used in MASS and MIPRO to each of the MAS we described above to make a comprehensive comparison. More details including the citations of the baselines are in Appendix-\ref{appendix:Baselines}. 

\begin{figure}[t]
	\centering
	\includegraphics[width=1\linewidth]{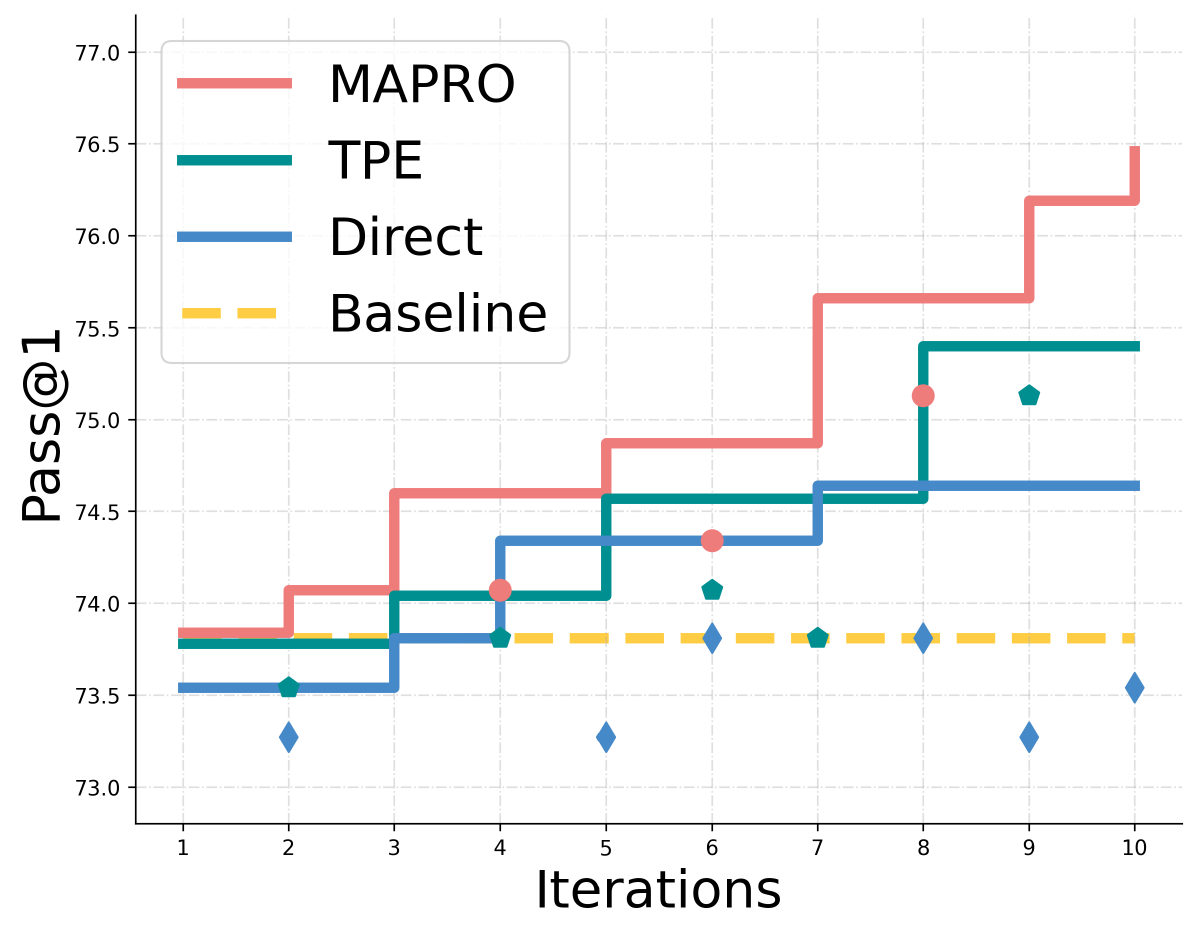}
        \vspace{-25pt}
	\caption{Optimization trajectories on the MBPP+ benchmark. We report the first ten optimization iterations using the chain MAS framework. MAPRO exhibits a more consistent and steady improvement compared to alternative methods.}
        \vspace{-20pt}
    \label{fig:optimization}
\end{figure}

\begin{table*}[t]
\centering
\resizebox{\textwidth}{!}{%
\begin{tabular}{c c c  c c c  c c  c c}
\toprule
\multicolumn{3}{c}{\textbf{Backbone: Llama 3.3-70b}} &
\multicolumn{3}{c}{\textbf{Code Generation}} &
\multicolumn{2}{c}{\textbf{Question Answering}} &
\multicolumn{2}{c}{\textbf{Math Reasoning}} \\
\cmidrule(lr){1-3}\cmidrule(lr){4-6}\cmidrule(lr){7-8}\cmidrule(lr){9-10}
\textbf{Model} & \textbf{MAS} & \textbf{Optimized} &
\textbf{HumanEval-ET} & \textbf{MBPP-Plus} & \textbf{CodeContest} &
\textbf{NewsQA} & \textbf{WebQuestion} &
\textbf{MATH} & \textbf{GSM8K} \\
\midrule
Raw            & \xmark & \xmark & 67.81 $\pm$ 0.86 & 68.04 $\pm$ 0.68 & 19.76 $\pm$ 2.08 & 58.65 $\pm$ 0.96 & 33.15 $\pm$ 0.50 & 67.56 $\pm$ 1.44 & 91.71 $\pm$ 0.37 \\
CoT            & \xmark & \xmark & 68.44 $\pm$ 1.31 & 68.04 $\pm$ 0.22 & 21.09 $\pm$ 1.31 & 60.56 $\pm$ 0.35 & 34.35 $\pm$ 0.47 & 69.01 $\pm$ 0.66 & 92.06 $\pm$ 0.29 \\
ReAct          & \xmark & \xmark & 69.06 $\pm$ 0.70 & 68.15 $\pm$ 0.44 & 20.36 $\pm$ 0.69 & 62.06 $\pm$ 0.28 & 35.84 $\pm$ 0.54 & 69.26 $\pm$ 0.84 & 92.34 $\pm$ 0.27 \\
\midrule
EvoPrompt      & \xmark & \cmark & 72.19 $\pm$ 1.71 & 69.74 $\pm$ 0.64 & 20.85 $\pm$ 1.33 & 64.37 $\pm$ 0.34 & 35.47 $\pm$ 0.41 & 71.62 $\pm$ 0.38 & 93.12 $\pm$ 0.19 \\
PromptBreeder   & \xmark & \cmark & 71.88 $\pm$ 1.10 & 69.10 $\pm$ 0.29 & 20.85 $\pm$ 1.40 & 64.53 $\pm$ 0.43 & 35.77 $\pm$ 0.16 & 71.01 $\pm$ 0.44 & 93.05 $\pm$ 0.23 \\
\midrule
Chain          & \cmark & \xmark & 70.00 $\pm$ 1.31 & 68.68 $\pm$ 0.14 & 27.52 $\pm$ 2.08 & 63.20 $\pm$ 0.45 & 35.24 $\pm$ 0.28 & 71.45 $\pm$ 1.04 & 93.71 $\pm$ 0.25 \\
w/t Direct     & \cmark & \cmark & 71.35 $\pm$ 1.80 & 69.58 $\pm$ 0.46 & 28.08 $\pm$ 0.70 & 63.62 $\pm$ 0.27 & 36.16 $\pm$ 0.18 & 70.37 $\pm$ 0.57 & 93.89 $\pm$ 0.31 \\
w/t TPE        & \cmark & \cmark & 71.88 $\pm$ 1.56 & 70.28 $\pm$ 0.40 & 28.69 $\pm$ 0.93 & 63.80 $\pm$ 0.21 & 36.04 $\pm$ 0.03 & 71.64 $\pm$ 0.62 & 93.42 $\pm$ 0.22 \\
\rowcolor{blue!8}
w/t MAPRO      & \cmark & \cmark & \best{75.00 $\pm$ 1.56} & \best{72.31 $\pm$ 0.55} & 30.10 $\pm$ 0.70 & 63.82 $\pm$ 0.54 & 36.22 $\pm$ 0.30 & 71.87 $\pm$ 0.35 & 93.56 $\pm$ 0.34 \\
\midrule
DMAD           & \cmark & \xmark & 70.94 $\pm$ 0.86 & 70.37 $\pm$ 0.46 & 34.06 $\pm$ 0.90 & 63.82 $\pm$ 0.54 & 35.56 $\pm$ 0.11 & 69.85 $\pm$ 0.42 & 94.12 $\pm$ 0.19 \\
w/t Direct     & \cmark & \cmark & 71.88 $\pm$ 1.56 & 70.90 $\pm$ 0.26 & 35.35 $\pm$ 0.70 & 64.12 $\pm$ 0.35 & 36.02 $\pm$ 0.20 & 70.99 $\pm$ 0.33 & \uline{94.93 $\pm$ 0.33} \\
w/t TPE        & \cmark & \cmark & 71.35 $\pm$ 1.80 & 70.55 $\pm$ 0.40 & 34.55 $\pm$ 0.61 & 64.30 $\pm$ 0.29 & 36.14 $\pm$ 0.22 & 71.32 $\pm$ 0.26 & 94.67 $\pm$ 0.28 \\
\rowcolor{blue!8}
w/t MAPRO      & \cmark & \cmark & 73.96 $\pm$ 0.90 & 71.60 $\pm$ 0.31 & 35.96 $\pm$ 1.53 & \uline{65.10 $\pm$ 0.24} & 36.22 $\pm$ 0.28 & \best{72.99 $\pm$ 0.33} & \best{95.91 $\pm$ 0.30} \\
\midrule
ChatEval (Swarm) & \cmark & \xmark & 71.25 $\pm$ 0.86 & 71.43 $\pm$ 0.19 & 34.91 $\pm$ 1.01 & 63.73 $\pm$ 0.43 & 36.44 $\pm$ 0.32 & 71.32 $\pm$ 0.26 & 93.14 $\pm$ 0.31 \\
w/t Direct     & \cmark & \cmark & 72.92 $\pm$ 0.90 & 70.99 $\pm$ 0.31 & 35.15 $\pm$ 0.61 & 64.02 $\pm$ 0.26 & 36.36 $\pm$ 0.27 & 71.73 $\pm$ 0.57 & 92.58 $\pm$ 0.37 \\
w/t TPE        & \cmark & \cmark & 72.40 $\pm$ 2.39 & 71.78 $\pm$ 0.31 & \uline{36.36 $\pm$ 0.61} & 64.18 $\pm$ 0.31 & \uline{36.41 $\pm$ 0.27} & 71.48 $\pm$ 0.80 & 92.45 $\pm$ 0.34 \\
\rowcolor{blue!8}
w/t MAPRO      & \cmark & \cmark & \best{75.00 $\pm$ 1.56} & \uline{72.22 $\pm$ 0.26} & \best{37.17 $\pm$ 0.93} & \best{65.45 $\pm$ 0.09} & \best{36.55 $\pm$ 0.29} & \uline{72.26 $\pm$ 0.71} & 92.38 $\pm$ 0.40 \\
\bottomrule
\end{tabular}%
}
\vspace{-7pt}
\caption{Performance results with baseline methods on Llama 3.3-70b. We report the mean and standard deviation for all results. The best performance is bolded and runner-ups are underlined. }
\vspace{-7pt}
\label{tab:result_llama}
\end{table*}

\begin{table*}[t]
\centering
\resizebox{\textwidth}{!}{%
\begin{tabular}{lccccccc}
\toprule
\textbf{Method} & \textbf{HumanEval-ET} & \textbf{MBPP-Plus} & \textbf{CodeContest} & \textbf{NewsQA} & \textbf{WebQuestion} & \textbf{MATH} & \textbf{GSM8K} \\
\midrule
MAPRO & 80.21 $\pm$ 0.90 & 76.54 $\pm$ 0.67 & 31.52 $\pm$ 0.61 & 64.00 $\pm$ 0.35 & 34.65 $\pm$ 0.30 & 64.30 $\pm$ 0.59 & 93.48 $\pm$ 0.42 \\
w/o demos & 76.04 $\pm$ 0.90 & 75.22 $\pm$ 0.31 & 29.70 $\pm$ 0.61 & 62.33 $\pm$ 0.31 & 34.20 $\pm$ 0.35 & 63.87 $\pm$ 0.23 & 92.86 $\pm$ 0.15 \\
\midrule
Drop (\%) & 5.20\% & 1.72\% & 5.78\% & 2.61\% & 1.30\% & 0.67\% & 0.66\% \\
\bottomrule
\end{tabular}%
}
\vspace{-7pt}
\caption{Ablation study results showing the performance drop when removing demonstration-guided reward. Numbers are reported with mean $\pm$ standard deviation, and relative drops are given in percentage.}
\vspace{-15pt}
\label{tab:ablation}
\end{table*}

\subsection{Main Results} 

We present the main results of MAPRO against baselines in Table~\ref{tab:result_haiku} and Table~\ref{tab:result_llama}. Across all benchmarks, MAPRO consistently achieves superior performance, often setting the best results within the same MAS, underscoring the strength of our approach. Several additional insights emerge. In terms of MAS structure, while topology exerts a stronger influence on overall accuracy than prompts, our plug-and-play design offers unmatched flexibility and extensibility, avoiding the heavy cost of topology optimization and enabling efficient deployment. As for task characteristics, MAPRO delivers the largest gains on reasoning-intensive tasks (e.g., WebQuestions, MBPP-Plus) compared to knowledge-heavy ones (e.g., NewsQA, CodeContest), highlighting the unique advantage of prompt optimization in complex reasoning. For LLM backbones, the results reaffirm general trends—Haiku excels in code, whereas Llama is stronger in reasoning—but also show that MAPRO adapts well across both. Notably, the optimal MAS under Llama shifted toward more sophisticated designs, suggesting that stronger reasoning models further amplify the benefits of our framework. Overall, these results validate MAPRO as both more effective and more versatile than existing methods, with significant potential for even greater gains on future LLMs.

\subsection{Optimization Trajectory}

We visualize the optimization trajectory of MAPRO as shown in Figure-\ref{fig:optimization}. MAPRO's trajectory demonstrates a more steady trend of optimization that gradually improves the validation performance towards better prompt sets, whereas we observe more fluctuations when it comes to other optimization methods, as they have a hard time capturing complicate interplays between agents. We further inspect an example of optimized prompt trajectory of an agent node in Table-\ref{tab:prompt-editing}. As can be seen, the prompt evolves overtime with more precise instructions that provides task-specific insights. These insights, especially the repeatedly occurring refinements, in practice, can be ingested into knowledge base which facilitates human-in-the-loop process and bring in more reliability and robustness to the system.

\subsection{Reward Model Analysis}

To assess the incremental gains of the reward models, we conducted an ablation study across all tasks. As shown in Table~\ref{tab:ablation}, the results underscore the critical role of demonstration-guided reward, consistent with TPO~\citep{li2025test}. A key insight is that the contribution of demonstrations varies across tasks, likely due to the relative simplicity of certain benchmarks such as those in the math domain. We also examined the consistency of the scoring process: under a low temperature setting, the selection procedure produced nearly identical outcomes across tasks and MAS configurations, so we omit ablation on that front. This robustness, together with the ablation results, demonstrate the efficacy of our reward model design.

\section{Conclusion}
We introduced MAPRO, a principled framework that first recasts multi-agent prompt optimization as a MAP inference problem and resolves it through language-guided belief propagation and topology-aware refinement. Across diverse downstream tasks, MAPRO consistently surpasses all types of baselines, demonstrating its effectiveness and generality. Beyond strong empirical gains, MAPRO delivers a plug-and-play setting that balances accuracy with flexibility. This flexibility, together with the interpretability provided by the optimization trajectories, making MAPRO especially practical for real-world MAS deployment and improvement.

\newpage
\section*{Limitations}
In this section, we discuss the limitations of our work and outline promising directions for future research. First, our study focuses exclusively on optimizing prompts for MASs while keeping the agent topology fixed. The results suggest that additional gains could be achieved through more deliberate choices of MAS topology. However, updating the topology requires reconfiguring the entire system, which is substantially more complex and resource-intensive. In many industrial applications where MAS designs are already deployed or constrained by fixed requirements, such re-establishment is impractical. This reflects a fundamental trade-off between flexibility, extensibility, efficiency, and performance. Nevertheless, extending MAPRO to jointly optimize both prompts and topologies would be an exciting avenue for future exploration. Second, while we employed LLM-based agents as reward models and demonstrated their efficacy and consistency, it would be valuable to investigate fine-tuned alternatives. In particular, approaches such as Max a Posteriori Policy Optimization \citep{abdolmaleki2018maximum} offer a principled framework that could replace our current reward mechanism and integrate more seamlessly into the overall optimization process. Exploring such directions could further enhance the robustness and generality of our approach.

\bibliography{reference}

\newpage
\appendix
\section{Related Work}
\label{appendix:Additional Related Work}
\subsection{Prompt Optimization for MAS}

Recent progress in large language models (LLMs) has enabled multi-agent systems (MAS), in which cooperating agents consistently outperform single agents on demanding reasoning and software-engineering tasks \citep{zhang2025agentrouter, zhang2025llms4all, zhong2024debug, wang2024executable}. This performance, however, depends on labor-intensive prompt engineering: each agent needs carefully crafted role instructions, and this effort grows rapidly as the number of agents increases, because aligning coordination between them becomes increasingly complex. To ease this burden, many studies now frame prompt design as an optimization problem and use heuristic search algorithms to explore and refine prompts with minimal human effort and oversight.

Prompt optimization in LLMs generally falls into two main categories: soft-prompt tuning in continuous space and discrete prompt optimization in text space. Soft tuning supports gradient-based updates but sacrifices transparency and portability, as its learned vectors are opaque, model-specific, and require gradient access that most black-box APIs do not provide \citep{cui2025automatic}. To work around this, researchers approximate gradients with LLM feedback and develop gradient-like strategies suited for non-differentiable settings. For example, some works apply beam search for step-wise refinement \citep{chen2024prompt, wangpromptagent}, while others explore alternative optimization strategies, such as evolutionary algorithms \citep{guo2025evoprompt, fernando2024promptbreeder} and other heuristic algorithms \citep{opsahl2024optimizing, li2025test}, to adapt prompts iteratively. Another related line of work focuses on prompt selection, searching a pool of variants to pick the best one \citep{wuprompt, wang2024one, song2025instantly}.

However, most existing methods target a single agent, while prompt optimization for MAS as a whole remains under-explored. Among the few efforts in this area, Mass \citep{zhou2025multi} warms each agent’s role prompt, prunes the interaction graph, and then jointly fine-tunes all prompts, showing that layered optimization boosts group performance. GPTSwarm \citep{zhuge2024gptswarm} treats agents as a graph and updates node-level prompts and edge connections together, letting prompts co-evolve with coordination patterns. We argue that the importance of proper prompt design has been significantly under-studied in these prior works. Specifically, current approaches still overlook two key issues: even minor lexical edits upstream can shift the distributions seen by downstream agents, and, to the best of our knowledge, none of the methods searches for a globally optimal set of prompts for the full system.

\subsection{LLM Agents for Code Generation Tasks}
Code generation has emerged as a core application for large language model (LLM) agents because it links natural-language reasoning with concrete, testable outputs and promises to automate sizable portions of software engineering and data-science workflows. Early studies tackled this task with single-agent or minimally interactive pipelines—such as Self-Refine \citep{madaan2023self}, Reflexion \citep{shinn2023reflexion}, and CoT-Zero \citep{kojima2022large}—that plan, execute, and iteratively repair their own code until unit tests pass. These works showed that even simple agent interactions can improve reliability when the agent can inspect failures and revise its output, a process that loosely aligns with causal reasoning \cite{ge2025review, jiang2024llm4causal}: the model infers potential sources of error and adjusts its generation in response.

As multi-agent systems advance, the trend has shifted toward complex MAS with carefully designed role-play interactions. Frameworks like CodeAct \citep{wang2024executable}, MetaGPT \citep{hong2024metagpt}, and ChatDev \citep{qian2024chatdev} assign specialized roles—planner, coder, tester, reviewer—and let agents converse in plain language, mimicking real software teams. These orchestrated exchanges boost division of labor and help solve coding problems that demand sophisticated reasoning, though they impose substantial overhead in prompt design. Building on these frameworks, recent work explores several directions. Some studies enhance individual modules through richer planning \citep{lei2025planning, chen2024divide}, stronger verification \citep{dainese2024generating, zheng2025makes}, or improved knowledge bases \citep{ouyang2025repograph, liu2025codexgraph}. Others introduce supervised signals into the framework, such as reinforcement learning \citep{feng2024natural}, to guide agent behavior. 

However, across all phases, prompt design remains a bottleneck: each role prompt must be carefully crafted, and even minor edits can ripple through the workflow. Consequently, a growing body of research now investigates automatic prompt optimization \citep{zhang2025g, zhou2025multi} to unlock more reliable and generalizable agent-collaboration schemes for real-world coding tasks.

\section{Implementation Details}
\subsection{Benchmarks}
\label{appendix:Benchmarks}
\textbf{HumanEval-ET}~\cite{dong2025codescore} is an extended benchmark for evaluating code generation. It builds upon the original HumanEval dataset by introducing more challenging variations and refined evaluation protocols, particularly emphasizing error tolerance and execution-based correctness. The dataset is specifically designed to better capture the robustness of large language models (LLMs) under real-world coding scenarios, where multiple correct implementations may exist and minor deviations from reference solutions should not necessarily be penalized. By incorporating these refinements, HumanEval-ET provides a more reliable and nuanced measure of code generation quality. Since this dataset doesn't provide a train-test split, we used the first 100 records for optimization and the rest 64 for zero-shot left-out testing.

\textbf{MBPP-Plus}~\citep{liu2023your} extends the "Mostly Basic Python Problems" (MBPP) dataset into a larger and more diverse collection. While MBPP was originally created to evaluate basic programming competency using short Python functions, MBPP-Plus expands both the scale and variety of tasks to cover more intricate programming constructs, edge cases, and multi-step logic. This augmentation addresses the limitations of the original dataset by providing a broader set of problems that better reflect practical coding challenges, thereby serving as a more comprehensive benchmark for evaluating code generation models.

\textbf{CodeContest}~\citep{li2022competition} is a benchmark derived from real competitive programming problems, representing a significant increase in difficulty compared to synthetic or basic coding datasets. It contains tasks sampled from programming competitions, where problems are designed to require algorithmic reasoning, data structure manipulation, and efficiency considerations. The inclusion of strict input–output constraints and hidden test cases makes CodeContest a rigorous benchmark that challenges LLMs to go beyond template-based solutions and demonstrate genuine problem-solving ability. Given the large volume of this dataset's training set, we sample the same records as the test set from training for optimization.

\textbf{NewsQA}~\citep{trischler2016newsqa} is a large-scale question answering dataset constructed from CNN news articles. It consists of over 100,000 human-generated questions paired with answers derived from corresponding news passages. Unlike earlier QA datasets that focus on simple fact extraction, NewsQA emphasizes reasoning, inference, and synthesis across multiple sentences within an article. Its design introduces ambiguity, unanswerable questions, and multi-sentence reasoning, making it a challenging benchmark for evaluating reading comprehension and open-domain question answering systems. Given the large volume of this dataset, we sample the first 500 records to use as optimization and left-out testing. 

\textbf{WebQuestions}~\citep{berant2013semantic} is a benchmark dataset for semantic parsing and knowledge-base question answering. It contains around 6,000 natural language questions paired with answers sourced from Freebase, covering a diverse range of topics. The dataset is notable for requiring models to bridge the gap between natural language queries and structured knowledge graph representations, thereby testing a system’s ability to perform entity linking, relation extraction, and logical reasoning. As one of the earliest large-scale QA datasets grounded in knowledge bases, WebQuestions has been widely adopted as a standard benchmark for semantic parsing and open-domain QA research. Given the large volume of this dataset, we sample the first 500 records to use as optimization and left-out testing. 

\textbf{MATH}~\citep{hendrycks2measuring} is a dataset specifically designed to evaluate advanced mathematical reasoning in LLMs. It contains approximately 12,000 competition-style problems, ranging from high school mathematics to Olympiad-level challenges, with step-by-step solutions provided. Unlike arithmetic-focused datasets, MATH covers a broad spectrum of topics including algebra, geometry, number theory, and calculus, requiring multi-step reasoning and symbolic manipulation. Its complexity makes it one of the most rigorous benchmarks for assessing the capacity of LLMs to handle formal reasoning and mathematical problem-solving. Given the large volume of this dataset's training set, we sample the same records as the test set from training for optimization.

\textbf{GSM8K}~\citep{cobbe2021training} (Grade School Math 8K) is a benchmark comprising 8.5k carefully crafted grade-school-level math word problems. Each problem is designed to require multi-step reasoning with arithmetic operations, testing a model’s ability to parse natural language descriptions, translate them into formal reasoning steps, and compute the correct answer. The dataset emphasizes chain-of-thought reasoning and has become a standard testbed for evaluating LLMs’ ability to perform reliable symbolic reasoning in relatively simple but compositional tasks. Its structured design and moderate difficulty level make GSM8K complementary to more advanced datasets like MATH. Given the large volume of this dataset's training set, we sample the same records as the test set from training for optimization.

\subsection{Baselines}
\label{appendix:Baselines}

\textbf{Chain-of-Thought (CoT)}~\citep{wei2022chain} is a prompting paradigm that encourages large language models (LLMs) to generate intermediate reasoning steps before arriving at final answers. Unlike direct-answer prompting, CoT exposes the model’s latent reasoning process, which has been shown to substantially improve performance on tasks requiring multi-step deduction such as arithmetic, commonsense inference, and symbolic reasoning. The introduction of CoT has established a new standard for eliciting reasoning from LLMs, making it a fundamental baseline in subsequent research. Its effectiveness also highlights a broader principle: structured prompting can significantly extend the reasoning capability of LLMs without the need for additional training.

\textbf{ReAct}~\citep{yao2023react} builds upon CoT by integrating reasoning with acting. Specifically, ReAct enables agents to interleave chain-of-thought reasoning with concrete actions, such as querying external knowledge sources, interacting with environments, or calling tools. This synergy allows models to dynamically refine their reasoning based on external feedback, thereby reducing hallucinations and improving factual grounding. ReAct has been validated across diverse tasks including knowledge-intensive QA, fact verification, and embodied agent settings, where its reasoning-and-acting paradigm consistently outperforms reasoning-only or acting-only strategies. As a baseline, ReAct represents an important step toward interactive and tool-augmented LLM systems.

\textbf{EvoPrompt}~\citep{guo2024connecting} frames prompt optimization as an evolutionary search process, where a population of prompts is iteratively mutated and recombined to generate stronger candidates. The method relies on large language models themselves as operators for variation, while selection mechanisms ensure gradual improvement. This makes EvoPrompt effective for black-box single-agent prompt optimization. However, its design remains confined to evolving isolated prompts, and it does not extend naturally to multi-agent settings where inter-agent coordination and topology play central roles.

\textbf{PromptBreeder}~\citep{fernando2024promptbreeder} extends evolutionary prompt optimization by introducing self-referential mutation. In this framework, not only task-prompts but also the mutation-prompts that generate them are evolved, enabling the system to adapt its own optimization strategy over time. This self-referential design yields a flexible and automated process for refining prompts in single-agent contexts. Nevertheless, PromptBreeder is inherently tailored to optimizing individual prompts and does not address the complexities of scaling to multi-agent systems.

\textbf{Chain}~\citep{shinn2023reflexion} represents the simplest combination topology of MAS. In our study, it combines the reasoning-and-acting paradigm of ReAct with a self-reflection module. After completing a task, the agent revisits its reasoning trajectory, identifies mistakes, and integrates corrective feedback into subsequent attempts. By incorporating the simple reflection module into the loop, Chain improves both robustness and sample efficiency. In our experiments, we adopt this variant as a baseline to capture the benefits of the effectiveness of simple MAS, compared with other MAS choices.

\textbf{DMAD}~\citep{liu2025breaking} (Diverse Multi-Agent Debate) is a recent state-of-the-art framework designed to overcome the inherent limitations of multi-agent debate (MAD). Traditional MAD setups often fall prey to a mental set, where agents—even if assigned different personas—rely on similar reasoning strategies, limiting their ability to explore alternative solutions. DMAD explicitly addresses this by requiring each agent to employ a distinct reasoning method (e.g., Chain-of-Thought, Step-Back Prompting, Program-of-Thought), thereby fostering genuine diversity in problem-solving. This represents an intermediate-complexity MAS topology that balances complexity and expressiveness. Given its robustness and state-of-the-art results, we consider DMAD an essential MAS base structure for evaluating MAS-level optimization.

\textbf{ChatEval}~\citep{chanchat2024eval} is a multi-agent evaluation framework that leverages structured dialogue among diverse LLM agents to produce more reliable judgments. In our study, we adopt the \textit{Simultaneous-Talk-with-Summarizer} variant, which we call SWARM, where agents contribute in parallel and a summarizer condenses their discussion into a concise shared history. In this setting, each agent interacts with each other in a dense format, making this baseline a typical and representative topology type, as it emphasizes on the richness of multi-agent discussion with strong reasoning and expressiveness.

\textbf{GPTSwarm}~\citep{zhuge2024gptswarm} frames language agents as computational graphs, where each node corresponds to an operation such as an LLM query, and edges capture the flow of information across agents. This framework is intended for optimizing both topology and prompts. To enable a fair comparison, we focus solely on the prompt optimization parts of this work. Within this framework, \textbf{Direct Optimization} is employed to refine the prompts associated with each node individually, using input–output histories and iterative updates to improve local performance. This strategy allows each operation to self-improve in isolation, but it treats prompts largely as independent units and does not account for the interdependencies across the wider agent graph. In contrast, our MAPRO framework explicitly models prompt optimization as a joint inference problem over the entire MAS topology, propagating credit and dependencies across nodes and edges. Thus, while GPTSwarm provides a strong formulation of node-level direct optimization, our approach generalizes this idea to coordinated optimization across multi-agent systems, addressing the limitations of local-only updates

\textbf{TPE Optimization}~\citep{zhou2025multi, opsahl2024optimizing} applies a Tree-structured Parzen Estimator (TPE)–based Bayesian search strategy to optimize prompts in multi-agent systems. In these frameworks, prompts (instructions and demonstrations) are treated as discrete parameters, and TPE is used to model the joint contribution of different parameter settings to downstream performance. This surrogate-based approach efficiently explores the search space by prioritizing promising configurations from past evaluations. Unlike \textbf{Direct Optimization}, which treats node prompts independently, TPE can partially capture dependencies between variables through its probabilistic modeling. However, TPE optimization still operates over a fixed pool of candidate proposals, limiting its ability to propagate credit across agents or adapt proposals dynamically. In contrast, our MAPRO framework formulates prompt optimization as a joint inference problem across the entire MAS topology, explicitly propagating dependencies and credit signals across nodes and edges. This enables coordinated optimization beyond the local or surrogate-based updates employed by TPE methods, addressing their limitations in capturing the full structure of multi-agent interactions.

\subsection{Training Protocol}
\label{appendix:training}

We limit the number of preference demonstrations to 3 and candidates to 5. We limit the agent number smaller than 10. We set model temperature at 0.2, maximum output tokens at 2048. We implement the same LLM backbone as both evaluator and executors in all phases. The optimized MAS is reported on the held-out test set over three runs, while other baselines over five runs. Given our mission to optimize the prompts, we didn't spend too much effort on prompt engineering, which mimics the real-life scenarios where a general prompt is adopted to a specific downstream tasks. The specific prompt designs can be seen in Appendix-\ref{appendix: prompts}. 

\section{Proof}
\label{appendix:proof}

\subsection{Proof of MAP Equivalence}
\label{appendix: proof map}

By Bayes’ rule, the classic MAP estimate chooses the hypothesis $P$ that maximizes the posterior:
\begin{equation}
\begin{aligned}
\widehat P_{\mathrm{MAP}}
&\in \arg\max_{P}\; \Pr(P\mid S) \\
&= \arg\max_{P}\; \Pr(S\mid P)\,\pi(P),
\end{aligned}
\end{equation}
where $S$ is the observed event and $\pi(P)$ is the prior on $P$. This way MAP finds the most probable explanation (the most likely hidden variable assignment) given what one observed.

In our case, $P=(p_1,\dots,p_N)\in P_1\times\cdots\times P_N$ is a joint prompt assignment and $S$ denote the event that the system succeeds on the batch. By construction of the node/edge success scores $g(\cdot),g(\cdot,\cdot)$,
\begin{equation}
\begin{aligned}
\Pr(S\mid P)
&= \prod_{i=1}^{N} \Pr(X_i{=}1\mid P)\ \!\!\!\!
   \prod_{(i,j)\in\mathcal{E}} \!\!\!\Pr(Y_{ij}{=}1\mid P) \\
&= \prod_{i=1}^{N} g(p_i)\; \!\!\!\!
   \prod_{(i,j)\in\mathcal{E}}\!\!\! g(p_i,p_j) \\
&=: \mathcal{T}(P).
\end{aligned}
\end{equation}

Since we do not assume one prompt set is inherently better than another (we have no prior knowledge). The most neutral choice is to use a uniform prior, and under a uniform prior, every $P\in P_1\times\cdots\times P_N$ is assigned the same 
positive probability. Thus $\pi(P)=c$ for some constant $c>0$ independent of $P$. 
Since multiplying by a constant does not affect an $\arg\max$, we have
\[
\arg\max_{P} f(P)\,c = \arg\max_{P} f(P).
\]

\noindent Therefore, given $\mathcal{T}(P)=\Pr(S\mid P)$,
\begin{equation}
\begin{aligned}
\arg\max_{P}\Pr(P\mid S)
&= \arg\max_{P}\Pr(S\mid P)\,\pi(P) \\
&= \arg\max_{P}\Pr(S\mid P)\,c \\
&= \arg\max_{P}\Pr(S\mid P) \\
&= \arg\max_{P}\mathcal{T}(P).
\end{aligned}
\end{equation}
Thus, maximizing the Joint Quality Score is exactly a MAP estimate of $P$. 

\subsection{Proof of Junction Tree MAP}
\label{appendix: proof junction tree}

Max-product belief propagation (MPBP) is guaranteed to compute the exact MAP
assignment only on tree-structured factor graphs. For a DAG $G=(\mathcal{V},\mathcal{E})$, the factorization ($\mathcal{T}(P)$) generally induces cycles, since a node $j$ with multiple parents couples the variables $\{p_i : (i,j)\in\mathcal{E}\}$ together. Formally, one first moralizes and triangulates the DAG to ensure a chordal structure admitting a junction tree.

The junction-tree construction converts this DAG factorization into an
equivalent tree-structured form. The procedure groups variables into clusters
$C \subseteq \mathcal{V}$, each associated with a potential $\psi_C$ defined as
\begin{equation}
\psi_C(P_C)
:= \prod_{i\in C} g(p_i)\; \!\!\!\!\!
   \prod_{\substack{(i,j)\in\mathcal{E}\\ \{i,j\}\subseteq C}}
   \!\!\! g(p_i,p_j),
\end{equation}
where $P_C=\{p_i : i\in C\}$. In words, every factor is assigned to
exactly one cluster that contains its variables. Clusters are arranged in a
tree $\mathcal{T}_{\text{JT}}$ satisfying the \emph{running intersection
property}: if a variable $p_i$ appears in two clusters $C_1,C_2$, then it
appears in every cluster on the unique path between $C_1$ and $C_2$ in
$\mathcal{T}_{\text{JT}}$.

The resulting representation is an exact refactorization:
\begin{equation}
\mathcal{T}(P)
= \prod_{C\in\mathcal{C}} \psi_C(P_C)\,
/ \prod_{s\in\mathcal{S}} \psi_s(P_s),
\end{equation}
where $\mathcal{S}$ denotes the separator sets (intersections of adjacent
clusters). Here each separator potential $\psi_s$ is defined as the product of factors assigned to $s$, ensuring no double counting. The division by separators ensures that no factor is double-counted and that the product reproduces exactly $\mathcal{T}(P)$. Since this is equivalent to the original joint score $\mathcal{T}(P)$ but expressed on a tree-structured factor graph, applying MPBP to $\{\psi_C\}$ on the junction tree yields
\begin{equation}
\arg\max_{P}\;\mathcal{T}(P)
= \arg\max_{P}\;
\frac{\prod_{C\in\mathcal{C}} \psi_C(P_C)}{\prod_{s\in\mathcal{S}} \psi_S(P_s)}.
\end{equation}
which recovers the exact MAP assignment of $P$. Hence, the junction-tree transformation converts a general DAG into a tree-structured model where MPBP can be applied directly and exactly. This ratio form follows from the junction-tree theorem, which guarantees that clique potentials multiplied and corrected by separator terms reproduce the exact joint distribution.

\subsection{Proof of MAP Global Optimality}
\label{appendix: proof global}

\emph{Lemma (optimal-subtree property).}
Assume the reward factorization is finite and has no negative factors, for any edge $i\!\to\! j$ and any $p_j$, $m_{i\to j}(p_j)$ equals the maximum of the product of factors contained in the subtree rooted at $i$, conditioned on $p_j$.

\emph{Proof.}
Assume $i$ is a leaf agent node, \eqref{eq:mpbp-message} reduces to $\max_{p_i} g(p_i)g(p_i,p_j)$, the best score of the leaf edge given $p_j$. Assume the claim holds for all children $k\in\text{\textit{Child}}(i)$.
Then the product inside \eqref{eq:mpbp-message} equals, for each fixed $p_i$, the optimal contributions of all child sub-trees consistent with $p_i$; maximizing over $p_i$ yields the optimal value of the entire subtree at $i$ given $p_j$.

\medskip
\emph{Theorem (global MAP optimality).}
Let
\(
p_r^*\in\arg\max_{p_r}\beta_r(p_r).
\)
Then there exists an assignment $P^\star$ obtained by the standard downward backtracking that satisfies
\(
P^*\in\arg\max_{P}\mathcal{T}(P),
\)

since the root collects optimal contributions from all disjoint subtrees. Thus
\[
\max_{p_r}\beta_r(p_r)=\max_{P}\mathcal{T}(P).
\]
During the upward pass, for every edge $i\!\to\! j$ and parent value $p_j$, the maximizer(s) achieving \eqref{eq:mpbp-message} define a witness choice $p_i^\star(p_j)$.
Starting from $p_r^*\in\arg\max\beta_r(p_r)$ and recursing $p_i^\star\!\left(p_j^*\right)$ along edges away from $r$ yields a full assignment $P^*$ that realizes the global maximum (ties broken arbitrarily).

\emph{Remark (junction tree).}
In the clique/sepset form, replace nodes $i$ by cliques $C$, parent $j$ by neighbor $D$, $p_i$ by $P_C$, and $g$ by $\psi$; messages are
\[
m_{C\to D}(P_{S_{CD}})\!\!=\!\!\max_{P_{C\setminus S_{CD}}}\!\!\bigl[\psi_C(P_C)\!\!\!\!\!\!\!\!\!\!\prod_{B\in\mathrm{nb}(C)\setminus\{D\}} \!\!\!\!\!\!\!\!\!\!m_{B\to C}(P_{S_{BC}})\bigr],
\]
and the same induction establishes exact MAP for the original DAG via the established equivalence.

\subsection{Time Complexity Analysis}
\label{appendix: time complexity}

\textbf{MPBP on a tree.}
Let $N=|\mathcal{V}|$ be the number of agents, $E=|\mathcal{E}|$ the number of edges,
and $K=\max_i |P_i|$ the maximum size of any agent’s prompt pool.
On a tree-structured factor graph, each edge passes one message in each direction.
Updating a single message requires comparing all prompt pairs $(p_i,p_j)$,
which costs $O(K^2)$.
Since there are $O(E)$ such messages in total, the overall complexity is
\[
O(EK^2).
\]

\noindent\textbf{Junction-Tree MAP (general DAG).}
For a DAG, we convert the graph into a junction tree whose clique set is denoted $\mathcal{C}$. Let $w$ be the induced treewidth, i.e.\ the size of the largest clique minus one. Each message update involves marginalizing/maximizing over a clique table of size $O(K^{w+1})$, and there are $O(|\mathcal{C}|)$ such cliques (at most linear in $N$).
Thus the complexity is
\[
O(|\mathcal{C}|\,K^{w+1}),
\]
with storage requirements of the same order.

The treewidth $w$ reflects how many agents must be grouped into a single clique
to remove cycles. For instance, if two agents both depend on the same parent, they may be merged into a summary clique of size three, so $w=2$. If a node has three parents, then a clique containing all four variables may be needed, giving $w=3$. In general, sparse MAS graphs usually have small $w$ (often $2$ or $3$), so the exponential factor $K^{w+1}$ remains modest. This means that in practical multi-agent settings, where each agent only interacts with a few neighbors, junction-tree MAP is efficient and scales nearly linearly with $N$ once $w$ is bounded.

In summary, we control both selection and update phase in polynomial time complexity and it scales well the increase of the density of interactions as well as the size of candidate pools, which emphasize the scalability and efficiency of our MAPRO framework.

\section{Prompt Designs}
\label{appendix: prompts}

\begin{figure*}[t]
\centering

\begin{tcolorbox}[colback=cadmiumgreen!5!white,colframe=cadmiumgreen!75!black, title=Node-Level Reward Model (Header + Prefix)]
\textbf{node\_header:}\\
You are a *reward model* for evaluating the competence, clarity of candidate **role prompts**.\\
Based on the input, output and prefernece examples,\\
you should first rank the candidate prompts with the good and bad examples,\\
Then you will give each a distinct two-decimal quality score between (0.00, 1.00) based on the standard and alignment with the good examples.\\
You should be severely harsh and the score difference should be ranged from 0.4 - 0.8 and each differs more than 0.05 with each other.\\
Finally, return exactly a score each line corresponding to the **prompt’s original position**. (Not the sorted score)\\
Note that your output should contain only the numeric scores (e.g., 0.62). Nothing else.\\[6pt]

\textbf{agent\_reward\_prefix:}\\
You are an evaluation LLM. Given \{input\} and the agent’s response \{output\}, rate how well the response accomplishes the agent’s role on a scale 0–1 (higher is better).Use the preference demonstrations below as reference.Return ONLY the floating-point score.\\[3pt]
=== Preference Demonstrations ===\\
\{demo\} \hfill \\
=== End Demonstrations ===
\end{tcolorbox}


\begin{tcolorbox}[colback=myred!5!white,colframe=myred!75!black, title=Edge-Level Reward Model (Header + Prefix)]
\textbf{edge\_header:}\\
You are a *reward model* for assessing **communication quality** from\\
an upstream agent to a downstream agent.  Consider information completeness, format,\\
clarity, and alignment with demonstrations.\\
Based on the input, output and prefernece examples,\\
you should first rank the candidate prompts with the good and bad examples,\\
Then you will give each a distinct two-decimal quality score between (0.00, 1.00) based on the standard and alignment with the good examples.\\
You should be severely harsh and the score difference should be ranged from 0.4 - 0.8 and each differs more than 0.05 with each other.\\
Finally, return exactly a score each line corresponding to the **prompt’s original position**. (Not the sorted score)\\
Note that your output should contain only the numeric scores (e.g., 0.62). Nothing else.\\[6pt]

\textbf{edge\_reward\_prefix:}\\
You are an evaluation LLM. Judge whether a message produced by agent \{i\} helps agent \{j\} perform its next step. Rate on a 0–1 scale.  Use the demonstrations for guidance. Return ONLY the floating-point score.\\[3pt]
=== Preference Demonstrations ===\\
\{demo\} \hfill  \\
=== End Demonstrations ===
\end{tcolorbox}
\vspace{-10pt}

\caption{Unified Reward Modeling Prompts for MAPRO: node-level (left) and edge-level (right), merging each module’s header and reward prefix verbatim.}
\label{fig:reward-models-merged}
\end{figure*}

\begin{figure*}[t]
\centering
\begin{tcolorbox}[colback=myblue!5!white,
                  colframe=myblue!75!black,
                  title= Feedback and Mutation Strategy Prompts]
\textbf{global\_feedback\_sys:}\\
You are an experienced prompt engineer and failure-analysis specialist.\\
Given multiple examples of runtime *error messages* produced by the given LLM-generated code,\\
identify the three most recurring but easy to solve root-cause patterns or missing constraints **in the prompts** that lead to the errors.  
Produce a short **specific and actionable** list of fix suggestions an author can apply.\\
Note 1: Output each fix as a bullet starting with numbers. Do NOT quote full stack traces; mention key function names only if essential.\\
Note 2: You should focus on the pragmatism and cleaniness of code rather than if it's easy to read, for example, if the a module doesn't have package `List`, 
instead of asking to properly import the package, you should emphasize it should write code without any type hints or annotations.\\[6pt]

\textbf{local\_feedback\_sys:}\\
You are a experienced prompt engineer and failure-analysis specialist. You are given: \\
1) The global overall feedback list that the system is currently facing.\\
2) Blame statements from downstream agents suggesting how the current module can be improved (may be empty).\\
3) The prompt this module is currently using.\\
Based on the roles of the current module, your task is to generate a *local feedback* list, focusing on give specific, actionable fix suggestions specifically for this current module to take to avoid downstream errors and satisfy the overall fix suggestions.  
Each line starts with ‘•’.\\[6pt]

\textbf{mutation\_strategy\_sys:}\\
You are a experienced prompt engineer and failure-analysis specialist.\\
You are given the original <prompt> of a module plus two feedback blocks:\\
One overall fix feedback suggesting the errors the system currently experience and one optional local feedback suggesting what this current modules can focus on to improve to benefit the system.\\
Your task is to modify, improve, and explode the original prompt by outputing exactly \{n\} JSON strings as prompt variations with specific and detailed improvement.\\
Note:\\
1) You should focus on the pragmatism and cleaniness of the prompts (You shouldn't acutally write any code), so **always emphasize** the code should be executable, wrapped in one function, without any type hints or annotations, and named as solution if no other names are provided.\\
2) You are only allowed to make relatively small edits. You must choose exactly one action item in the following:  
a) adding one sentence from the feedback.  
b) replacing one senetence from the feedback to existing edits.  
c) Re-organize, rewrite or clean the current prompt to make it logically consistent.  
d) delete one redundant sentence in the current prompt.\\
3) You should ALWAYS respond with ONLY the VALID JSON array – You should return No headings, no prose such as </prompt>, no markdown fences such as ```, no trailing commas, no escape codes, or unclosed parenthesis.  
Each string must be valid UTF-8. Escape all newlines as \textbackslash n. No raw newlines inside JSON strings.  
Example (node, n = 2): ["Prompt variant 1","Prompt variant 2"].
\end{tcolorbox}
\vspace{-10pt}
\caption{Feedback system prompts in MAPRO (for coding tasks): global feedback, local feedback, and mutation strategy.}
\label{fig:feedback-prompts}
\end{figure*}

\begin{figure*}[t]
\centering

\begin{tcolorbox}[colback=cadmiumgreen!5!white,
                  colframe=cadmiumgreen!75!black,
                  title=Variation Prompt]
\textbf{variation:}\\
You are a prompt-engineering assistant.\\
The user will give you an original prompt TEMPLATE inside <prompt></prompt>.\\
Produce \{n\} diverse textual prompt variants (NOT solution, but the prompts) that keep the same intent but differ in wording, ordering, or tone.  
Note that you should generate the prompt for the agent not generate solution.\\
Don't write code here and Return **only** a JSON array of strings.\\
Respond on a single line only. Do not emit any raw line breaks.
\end{tcolorbox}

\begin{tcolorbox}[colback=myred!5!white,
                  colframe=myred!75!black,
                  title=Negative Variation Prompt]
\textbf{neg\_variation:}\\
You are a prompt-mutation helper.\\
The user will give you a JSON object with: \\
good\_examples : list[str]        \# 3 GOOD prompt templates (node) *or* 3 GOOD upstream-downstream pairs\\
mode          : "node"|"edge"    \# mutation type\\
n             : int              \# number of BAD variants requested\\
Produce exactly \{n\} sligthly BAD variants: \\
• For "node": each string could omit some key instructions, introduce contradictions, or add irrelevant text that reduces agent quality.\\
• For "edge": each string code be a JSON array ["bad\_upstream", "good\_downstream"] where bad\_upstream makes the pair incompatible.\\
• Note that your generation should be obviously worse than good examples, but not too absurd or entirely off the topic.\\
Remember, Return nothing except one valid JSON array.\\
- For mode = "node" → ["str", "str", …]\\
- For mode = "edge" → [["str","str"], ["str","str"], …]\\
You should ALWAYS respond with ONLY the VALID JSON array – You should return No headings, no prose such as </prompt>, no markdown fences such as ```, no trailing commas, no escape codes, or unclosed parenthesis.\\
Each string must be valid UTF-8. Escape all newlines as \textbackslash n. No raw newlines inside JSON strings.
\end{tcolorbox}

\vspace{-10pt}
\caption{Initialization prompts in MAPRO: \textit{variation} (left) for diverse positive variants and \textit{neg\_variation} (right) for intentionally degraded variants.}
\label{fig:initialization-prompts}
\end{figure*}

\begin{figure*}[t]
\centering
\begin{tcolorbox}[colback=myblue!5!white,
                  colframe=myblue!75!black,
                  title=Coding Prompts and Notes ]

\textbf{raw:}\\
You are a reasoning agent and coding expert. Solve the task by outputting \textbf{only executable Python code} as a \textbf{single function}. 
Do not print any prose, comments, or markdown fences. If preprocessing or postprocessing is needed to conform to the specified input/output format,
perform it \emph{inside} the function. Follow all constraints in \textbf{note} exactly.

\vspace{6pt}
\textbf{cot:}\\
You are a reasoning agent and coding expert. First reason step by step \emph{silently} (do not print thoughts or a plan). 
Then output \textbf{only executable Python code} as a \textbf{single function} that solves the task with the correct input/output format.
Do not include comments, explanations, tags, or markdown fences. Follow \textbf{note} exactly.

\vspace{6pt}
\textbf{react:}\\
You are a reasoning agent. First analyze the problem and form a plan \emph{silently} (do not print it). 
Then use that plan to produce the final answer as \textbf{code only} — a \textbf{single Python function} with no comments, prose, or markdown fences. 
If \texttt{FEEDBACK} is provided, revise the code \emph{only} when the feedback is correct and improves conformance to the task or \textbf{note};
otherwise keep the best previous solution. Follow \textbf{note} exactly.

\vspace{6pt}
\textbf{reflect:}\\
You are a coding \textbf{critic}. Be conservative — revise only if there are concrete mistakes. Evaluate the submitted answer against:\\
1) It is \textbf{only} executable Python code with \textbf{no} markdown, tags, or comments, and has no obvious syntax errors.\\
2) It implements any required \textbf{preprocessing/postprocessing} so the input and output formats are correct.\\
3) It satisfies all constraints in \textbf{note}.\\
If everything is correct, reply \textbf{exactly}: \texttt{ACCEPT}\\
Otherwise, reply \textbf{exactly}: \texttt{REVISE: <concise, actionable fixes required>}\\
Do not include code blocks, bullets, or extra text beyond the required format.

\vspace{6pt}
\textbf{note:}\\
1) \textbf{No type annotations or return-type hints}.\\
2) \textbf{Output only executable Python code}, with no tags (e.g., </...>), no markdown fences (```), and no explanations or comments.\\
3) \textbf{Wrap the solution in exactly one function}. If the function name is specified in the problem or PUBLIC TEST, use it and keep the exact parameter list. 
Otherwise, name the function \texttt{solution}.\\
4) \textbf{Match the task’s input format}. If examples indicate inputs arrive as strings, accept a string parameter and parse internally. 
Likewise, format outputs exactly as required (e.g., print vs return).\\
5) Use only Python’s standard library; do \textbf{not} rely on network access, external files, or third-party packages.\\
6) Ensure deterministic behavior and avoid unnecessary randomness or system calls.\\
7) Prefer clear, robust algorithms; handle edge cases implied by the task (empty inputs, boundary values) when reasonable.

\end{tcolorbox}

\vspace{-10pt}
\caption{Unified coding prompt suite for coding solutions. The five roles (RAW, CoT, ReAct, Reflect, Note) ensure silent planning, strict code-only output, conservative review, and precise conformance to I/O constraints.}
\label{fig:coding-prompts}
\end{figure*}

\begin{figure*}[t]
\centering
\begin{tcolorbox}[colback=myred!5!white,
                  colframe=myred!75!black,
                  title=Math Prompts and Notes]

\textbf{raw:}\\
You are a competition mathematician. Solve the problem with clear step-by-step reasoning, using exact symbolic forms (prefer fractions to decimals when appropriate). 
End with a single final line: \(\boxed{\langle answer\rangle}\). Follow \textbf{note} exactly.

\vspace{6pt}
\textbf{cot:}\\
First think step by step \emph{silently} (do not print your thoughts); then present a concise, logically ordered solution. 
Use exact forms; avoid decimals unless requested. End with the single final line: \(\boxed{\langle answer\rangle}\).
Follow \textbf{note}.

\vspace{6pt}
\textbf{react:}\\
Analyze the problem and form a plan \emph{silently} (do not print it); then present a full worked solution with clear steps and exact forms. 
If \texttt{FEEDBACK} is provided, revise the solution only when it is correct and improves adherence to \textbf{note}; otherwise keep the best prior solution. 
Finish with the final line: \(\boxed{\langle answer\rangle}\).

\vspace{6pt}
\textbf{reflect:}\\
You are a competition-math \textbf{critic}. Assess the submitted solution on:\\
1) Mathematical correctness of the result and reasoning.\\
2) Output formatting (final line exactly boxed).\\
3) Adherence to all constraints in \textbf{note}.\\
If everything is correct, reply \textbf{exactly}: \texttt{ACCEPT}\\
Otherwise, reply \textbf{exactly}: \texttt{REVISE: <concise, actionable fixes required>}\\
Do not include any extra text beyond the required format.

\vspace{6pt}
\textbf{note:}\\
\textit{MATH Canonicalization (Top Priority)}\\
1) Final line only: \(\boxed{\langle answer\rangle}\).\\
2) Exact forms: reduce \(a/b\); simplify radicals; use \(\pi\); avoid decimals unless asked.\\
3) Numbers: no commas anywhere (9901 not 9,901; \(448/15625\) not \(2{,}240/78{,}125\)).\\
4) Expressions: canonical, no spaces, use \verb|^| for exponents (\(x^3+3x-6\)). Do not prepend variables or '=' (prefer \(5\) over \(x=5\)).\\
5) MCQ: box the single capital letter only.\\
6) Tuples/Sets: \((a,b,\dots)\) and \(\{a,b,\dots\}\) with simplified components.\\
7) Units: match the problem; use \(k^\circ\) for degrees; default radians.\\
8) Sanity: respect domains; drop extraneous roots; include a quick plug-back check.

\end{tcolorbox}

\vspace{-15pt}
\caption{Unified math prompt suite for exact, well-formatted solutions. The five roles (RAW, CoT, ReAct, Reflect, Note) enforce silent planning, precise symbolic work, and a canonical boxed final answer. Different from coding prompts, prompts for question answering are rather similar to math prompts as they both relate to reasoning, thus we skip the demonstration here.}
\label{fig:math-prompts}
\end{figure*}

\end{document}